\documentclass[10pt,doublecolumn]{IEEEtran}
\usepackage{amsmath}
\usepackage{amssymb}
\usepackage{amsthm}
\usepackage{epsfig}

\usepackage{graphics}

\renewcommand{\vec}[1]{\ensuremath{{\rm\bf{#1}}}}
\newcommand{\mat}[1]{\ensuremath{{\rm\bf{#1}}}}

\newtheorem{theorem}{Theorem}

\ifCLASSINFOpdf
\else
\fi

\begin{document}
\title{Bayesian Model Selection Methods for Mutual and Symmetric $k$-Nearest Neighbor Classification}

\author{Hyun-Chul Kim
\thanks{H.-C. Kim is with R$^2$ Research, 
       Seoul, South Korea.    \newline
        E-mail: hckim.sr{\fontfamily{ptm}\selectfont @}gmail.com}  } 

\maketitle

\begin{abstract}
The $k$-nearest neighbor  classification method  ($k$-NNC) is one of the simplest nonparametric classification methods. The mutual $k$-NN classification method (M$k$NNC) is a variant of $k$-NNC based on mutual neighborship.  We propose another variant of $k$-NNC, the symmetric $k$-NN classification method (S$k$NNC) based on both mutual neighborship and one-sided neighborship.  The performance of M$k$NNC and S$k$NNC depends on the parameter $k$ as the one of $k$-NNC does. We propose the ways how M$k$NN and S$k$NN classification can be performed based on Bayesian mutual and symmetric $k$-NN regression methods with the selection schemes for the parameter $k$.  Bayesian mutual and symmetric $k$-NN regression methods are based on Gaussian process models, and it turns out that they can do M$k$NN and S$k$NN classification with new encodings of target values (class labels).  The simulation results show that the proposed methods are better than or comparable to $k$-NNC, M$k$NNC and S$k$NNC with the parameter $k$ selected by the leave-one-out cross validation method not only for an artificial data set but also for real world data sets.
\end{abstract}


% \fi
\IEEEpeerreviewmaketitle

\section{Introduction}

One of the well-known nonparametric classification methods is  $k$-nearest neighbor ($k$-NN) classifiation method \cite{fix1951discriminatory,fix1952discriminatory,cover1967nearest}. It uses one of the simplest rules among nonparametric classification methods.  It assigns to a given test data point the most frequent class label appearing in the set of $k$ nearest data points  to the test data point.  Performance of $k$-NN classifiers is influenced by the distance measure \cite{short1981optimal} and the parameter $k$\footnote{More accurately $k$ should be called the hyperparameter since $k$-NN is a nonparametric method, but in this paper we also call it the parameter according to the convention.} \cite{Holmes_ProbkNNpr,ghosh2006optimum}.  So it is an important issue to select the best distance mesure and the best parameter $k$ in $k$-NN classification. In this paper we focus on the selection of the best parameter $k$ in the variates of $k$-NN classification although the selection of the best distance measure is also important.

\cite{Holmes_ProbkNNpr} proposed an approximate Bayesian approach to $k$-NN classification, where a single parameter $k$ was not selected but its posterior distribution was estimated. It was not exactly probabilistic because of the missing of the proper normalization constant in the model as mentioned in \cite{Cucala08BayesiankNNcla}. It provided the class probability for a test data point and the approximate distribution of the parameter $k$ by MCMC methods.  It was followed by an alternative model with likelihood-based inference and a method to select the best parameter $k$  based on BIC (Bayesian information criterion) method \cite{holmes2003likelihood}.  While those models are not fully probabilistic, \cite{Cucala08BayesiankNNcla} proposed  a full
Bayesian probabilistic model for $k$-NN classification based on a symmetrized Boltzmann modelling with various kinds of sampling methods including a perfect sampling.  Due to the symmetrized modification, their model does not fully reflect $k$-NN classification any more (e.g. it does not have asymmetry such as the one in $k$-NN classification.).  So the most probable $k$ in their model may not be optimal in $k$-NN classification.

\cite{ghosh2006optimum} has proposed another method to select the optimal parameter $k$ based on approximate Bayes risk. They modelled class probabilities for each training data point based on  $k$-NN density estimation in the leave-one-out manner.  Starting from those class probabilities by applying Bayes' rule they get the accuracy index $\alpha(k)$. They proved that $1-\alpha(k)$ asymptotically converges to the optimal Bayes risk. Their simulation results showed that their proposed methods were better than cross-validation and likelihood cross-validation techniques.

Mutual $k$-NN (M$k$NN) classification is a variate of $k$-NN classfication based on mutual neighborship rather than one-sided neighborship.  M$k$NN concept was applied to clustering tasks \cite{gowda1978agglomerative,gowda1979condensed}. More recently, M$k$NN methods have been applied to classification \cite{liu2010new}, outlier detection \cite{hautamaki2004outlier}, object retrieval \cite{jegou2010accurate}, clustering of interval-valued symbolic patterns \cite{guru2006clustering}, and regression \cite{JMLR:v14:guyader13a}.  \cite{ozaki2011using} used M$k$NN concept to semi-supervised classification of natural language data and showed that the case of
using M$k$NN concept consistently outperform the case of using $k$-NN concept.

We propose another variate of $k$-NN classification, symmetric $k$-NN (S$k$NN) classification motivated by a symmetrized modelling used in \cite{Cucala08BayesiankNNcla}. S$k$NN consider neighbors both with mutual neighborship and one-sided neighborship. In S$k$NN classification, one-sided neighbors contribute to the decision in the same way as in $k$-NN classification, and mutual $k$-nearest neighbors contribute to the deicision twice more than one-sided $k$ nearest neighbors.

We propose Bayesian methods to select the parameter $k$ for M$k$NN\footnote{To our knowledge no Bayesian model selection method for M$k$NN classiifcation has been proposed.} and S$k$NN classification. 
This paper does not propose Bayesian probabilistic models for M$k$NN and S$k$NN classification, but model selection methods for them in the Bayesian evidence framework are proposed.  The methods are based on Bayesian M$k$NN and S$k$NN regression methods, with which M$k$NN and S$k$NN classification can be done.  A model selection method for S$k$NN classification is related to the estimation of the parameter $k$ in \cite{Cucala08BayesiankNNcla}, because the model proposed in \cite{Cucala08BayesiankNNcla} can be regarded a Bayesian probabilistic model for S$k$NN classification. 

The paper is organized as follows.  In section \ref{sec:Bayes_mut_sym_nnr} we describe mutual
$k$-NN and symmetric $k$-NN regression, and their Bayesian extensions with the selection method for the parameter $k$.  In Section \ref{sec:mut_sym_knn_cla} we explain how we can do M$k$NN and S$k$NN classification with Bayesian M$k$NN and S$k$NN regression methods with their own selection schemes for the  parameter $k$.
In section \ref{sec:sim_res} we show simulation results for an artificial data set and real-word data sets. Finally a conclusion is drawn.

\section{Bayesian Mutual and Symmetric $k$-Nearest Neighbor Regression}

\label{sec:Bayes_mut_sym_nnr}

\subsection{Mutual and Symmetric $k$-Nearest Neighbor Regression}

\label{subsubec:mut_sym_nnr}

Let $\mathcal{N}_k(\vec x)$ be the set of the $k$ nearest neighbors of $\vec x$ in $\mathcal{D}_n$, $\mathcal{N}'_k(\vec x_i)$ the set of $k$ nearest neighbors of $\vec x_i$ in $(\mathcal{D}_n \backslash \{\vec x_i \}) \cup \{ \vec x \}.$ The set of mutual nearest neighbors of $\vec x$ is defined as
\begin{align}
	\mathcal{M}_k(\vec x) = \{ \vec x_i \in \mathcal{N}_k(\vec x) : \vec x \in \mathcal{N}'_k (\vec x_i) \}.
\end{align}
Then, the mutual $k$-nearest neighbor regression estimate is defined as
\begin{align}
    m^{\mbox{M$k$NNR}}_n(\vec x) = \left\{ \begin{array}{ll}
         \frac{1}{M_k(\vec x)}\sum_{i:\vec x_i \in \mathcal{M}_k(\vec x)}^k y_i & \mbox{if $M_k(\vec x) \neq 0$};\\
        0 & \mbox{if $M_k(\vec x) = 0$}.\end{array} \right.
\label{eqn:mut_nn_reg_est}
\end{align}
where $M_k(\vec x) = | \mathcal{M}_k(\vec x)| $  \cite{JMLR:v14:guyader13a}. 

Motivated by the symmetrised modelling for the $k$-NN classification in \cite{Cucala08BayesiankNNcla}, we define the symmetric $k$-nearest neighbor regression estimate as
\begin{align}
    &m^{\mbox{S$k$NNR}}_n(\vec x) \nonumber \\ 
       &= \frac{1}{N_k(\vec x)+N'_k(\vec x)} 
         \sum_{i:\vec x_i \in \mathcal{N}_k(\vec x)}^k  ( \delta_{\vec x_i \in \mathcal{N}_k (\vec x)} + \delta_{\vec x \in \mathcal{N}'_k (\vec x_i)} ) y_i.
\label{eqn:sym_nn_reg_est}
\end{align}
where $N_k(\vec x) = | \mathcal{N}_k(\vec x)| $ and $N'_k(\vec x) = | \mathcal{N}'_k(\vec x)|$.

\subsection{Bayesian Mutual and Symmetric $k$-NN Regression via Gaussian Processes}
\label{subsec:Bayes_mut_sym_nnr}

\subsubsection{Gaussian Process Regression}

Assume that we have a data set $D$ of data points $\vec x_i$ with
continuous target values $y_i$: $D = \{(\vec x_i, y_i)|i = 1, 2,
\ldots, n\}$, $X = \{\vec x_i|i = 1, 2, \ldots, n \}$, $\vec y=[y_1,
y_2, \ldots, y_n]^T)$. We assume that the observations of target values are nosiy,
and set $y_i=f(\vec x_i)+\epsilon_i$, where $f(\cdot)$ is a target function to be estimated
and $\epsilon_i \sim \mathcal{N}(0,v_1)$. A function $f(\cdot)$ to be estimated given
$D$ is assumed to have Gaussian process prior, which means that any
collection of functional values are assumed to be multivariate
Gaussian.

The prior for the function values $\vec f(=[f(\vec x_1) f(\vec x_2) \ldots f(\vec x_n) ]^T)$ is assumed to be Gaussian:
\begin{align}
p(\vec f|X, \Theta_f) = \mathcal{N}(\vec 0, \mat C_f).
\end{align}
Then the density function
for the target values can be described as follows.
\begin{align}
	p(\vec y| X, \Theta) &= \mathcal{N}(\vec 0, \mat C_f + v_1 \mat I) \\
					&= \mathcal{N}(\vec 0, \mat C),
\end{align} 
where $\mat C$ is a matrix whose elements $C_{ij}$ is a covariance
function value $c(\vec x_i,\vec x_j)$ of $\vec x_i$, $\vec x_j$ and
$\Theta$ is the set of hyperparameters in the covariance function. 

It can be shown that GPR provides the following
distribution of target value $f_{\mathrm{new}}(=f(\vec x_{\mathrm{new}}))$ given a
test data $\vec x_{\mathrm{new}}$:
\begin{align}
 p(f_{\mathrm{new}}|\vec x_{\mathrm{new}}, D,\Theta)=\mathcal{N}(\vec k^T \mat
 C^{-1} \vec f, \kappa - \vec k^T \mat C^{-1} \vec k),
 \label{eqn:result}
\end{align}
where $\vec k = [c(\vec x_{\mathrm{new}},\vec x_1) \ldots c(\vec
x_{\mathrm{new}},\vec x_n)]^T$, $\kappa=c(\vec x_{\mathrm{new}},\vec x_{\mathrm{new}})$. The
variance of the target value $f_{\mathrm{new}}$ is related to the degree of
its uncertainty.  We can select the proper $\Theta$ by
maximizing the marginal likelihood $p(\vec y|X,\Theta)$
\cite{Williams95gaussian,Gibbs97efficientimplementation,Rasmussen:2005:GPM:1162254},
or we can average
over the hyperparameters with MCMC methods \cite{Williams95gaussian,Neal1997RC_GP}.

\subsubsection{Laplacian-based Covariance Matrix}

The combinatorial Laplacian $\mat L$ is defined as follows.
\begin{align}
\mat L = \mat D - \mat W,\label{eq:Delta}
\end{align}where $\mat W$ is an $N\times N$ edge-weight matrix with the edge
weight between two points $\vec x_i$,$\vec x_j$ given as $w_{ij}(=w(\vec x_i,\vec x_j))$ and $\mat D =
\mathrm{diag}(d_1,...,d_N)$ is a diagonal matrix with diagonal
entries $d_i=\sum_{j} w_{ij}$.

Similarly to \cite{zhu2003semi}, to avoid the singularity we set Laplacian-based covariance matrix as
\begin{align}
   \mat C =&  (\mat L + \sigma^2 \mat I)^{-1} = \tilde{\mat C}^{-1}.
\end{align}
Then, we have Gaussian process prior as follows.
\begin{align}
	p(\vec y| X, \Theta) &= \mathcal{N} (\vec 0, \mat C),
\end{align}
The predictive distiribution for $y_{\mathrm{new}}$ is as follows (See \cite{kim2016Bayes_kr} for the detailed derivation).
\begin{align}
	p(y_{\mathrm{new}}| \vec y, X, \vec x_{\mathrm{new}}, \Theta) & \propto  \mathcal{N}(-\frac{1}{\tilde{\kappa}} \tilde{\vec k}^T \vec
 y,\frac{1}{\tilde{\kappa}}),
\end{align}
where
\begin{align}
   \tilde{\kappa} &= \sum_{i=1}^N w(\vec x_{\mathrm{new}},\vec x_i) + \sigma^2, \\
   \tilde{ \vec k}^T &= - [ w(\vec x_{\mathrm{new}}, \vec x_1), w(\vec x_{\mathrm{new}}, \vec x_2), \ldots, w(\vec x_{\mathrm{new}}, \vec x_N) ].
\end{align}
The mean and variance of $y_{\mathrm{new}}$ is represented as
\begin{align}
   \mu_{ y_{\mathrm{new}}}&= -\frac{1}{\tilde{\kappa}} \tilde{\vec k}^T \vec y_L 
                            =\frac{\sum_{i=1}^N w (\vec x_{\mathrm{new}}, \vec x_i) y_i }{\sum_{i=1}^N w (\vec x_\mathrm{new}, \vec x_i)+\sigma^2}, \label{eq:transd_lap_mean} \\
   \sigma^2_{ y_{\mathrm{new}}}&= \frac{1}{\tilde{\kappa}}
                            =\frac{1}{\sum_{i=1}^N w(\vec x_{\mathrm{new}},\vec x_i) + \sigma^2}. \label{eq:transd_lap_std}
\end{align}

\subsubsection{Bayesian Mutual and Symmetric $k$-NN Regression}

First, we describe Bayesian mutual $k$-NN regression proposed in \cite{kim2016Bayes_kr}.
When we replace $w_{ij}(=w(\vec x_i,\vec x_j))$ with the function
\begin{align}
  w_{\mathrm{M}\it{k}\mathrm{NN}}(\vec x_i,\vec x_j) =  \sigma_0  \delta_{\vec x_j \sim_k \vec x_i} \cdot \delta_{\vec x_i \sim_k \vec x_j},
\end{align}
where the relation $\sim_k$ is defined as
\begin{align}
	\vec x_i \sim_k \vec x_j = \left\{ \begin{array}{ll}
 T  & \mbox{if $j \neq i$ and $\vec x_j$ is a $k$-nearest neighbor} \\
    & \mbox{of $\vec x_i$}; \\
   F   & \mbox{otherwise,} \end{array} \right.
\end{align}
and apply Eq (\ref{eq:transd_lap_std}), we get Bayesian mutual $k$-NN regression estimate given a new data $\vec x_{\mathrm{new}}$ as follows.
\begin{align}
   m^{\mbox{BM}k\mbox{NNR}}_n(\vec x_{\mathrm{new}}) =    	\mu_{f_{\mathrm{new}},\mathrm{M}\it{k}\mathrm{NN}},
\end{align}
where
\begin{align}
   \mu_{f_{\mathrm{new}},\mathrm{M}\it{k}\mathrm{NN}}
=&\frac{\sum_{i=1}^N w_{\mathrm{M}\it{k}\mathrm{NN}} (\vec x_{\mathrm{new}}, \vec x_i) y_i }
{\sum_{i=1}^N w_{\mathrm{M}\it{k}\mathrm{NN}} (\vec x_{\mathrm{new}}, \vec x_i)+\sigma^2} \nonumber \\
 =&\frac{\sum_{i=1}^N  \delta_{\vec x_j \sim_k \vec x_i} \cdot \delta_{\vec x_i \sim_k \vec x_j} y_i }
{\sum_{i=1}^N \delta_{\vec x_j \sim_k \vec x_i} \cdot \delta_{\vec x_i \sim_k \vec x_j} +\sigma^2/\sigma_0 }.
\label{eqn:Bayes_mut_knn_reg}
\end{align}
We have two following theorems about the validity of the covariance matrix with $w_{\mathrm{M}\it{k}\mathrm{NN}}(\vec x_i,\vec x_j)$ and asymptotic property of the regression estimate. (See \cite{kim2016Bayes_kr} for the proofs.)
\begin{theorem}
   Covairance matrix $\tilde{\mat C}$ with $w_{ij}(=w_{\mathrm{M}\it{k}\mathrm{NN}}(\vec x_i,\vec x_j))$ is valid for Gaussian processes if $\sigma^2>0$.
\end{theorem}
\begin{theorem}
  $\mu_{f_{\mathrm{new}},{\mathrm{M}{\it k}\mathrm{NN}}} (=-\frac{1}{\tilde{\kappa}} \tilde{\vec k}^T \vec y)$ converges to mutual $k$-$\mathrm{NN}$ regression as $\sigma^2/\sigma_0$ approaches $0$.
\end{theorem}

Now related to symmetric $k$-NN regression, we propose Bayesian symmetric $k$-NN regression. 
\begin{align}
  w_{\mathrm{S}\it{k}\mathrm{NN}}(\vec x_i,\vec x_j) = \sigma_0 ( \delta_{\vec x_j \sim_k \vec x_i}  + \delta_{\vec x_i \sim_k \vec x_j}   ) 
\end{align}
Similarly to Eq (\ref{eq:transd_lap_mean}) Bayesian symmetric $k$-NN regression is obtained as follows.
\begin{align}
   m^{\mbox{BM}k\mbox{NNR}}_n(\vec x_{\mathrm{new}}) =    	\mu_{f_{\mathrm{new}},\mathrm{M}\it{k}\mathrm{NN}},
\end{align}
where
\begin{align}
   \mu_{f_{\mathrm{new}},\mathrm{S}\it{k}\mathrm{NN}}
=& \frac{\sum_{i=1}^N w_{\mathrm{S}\it{k}\mathrm{NN}} (\vec x, \vec x_i) y_i }
{\sum_{i=1}^N w_{\mathrm{S}\it{k}\mathrm{NN}} (\vec x, \vec x_i)+\sigma^2} \\
=& \frac{\sum_{i=1}^N   (\delta_{\vec x_j \sim_k \vec x_i} +\delta_{\vec x_i \sim_k \vec x_j} ) y_i }
{\sum_{i=1}^N ( \delta_{\vec x_j \sim_k \vec x_i} +\delta_{\vec x_i \sim_k \vec x_j}) +\sigma^2/\sigma_0 }.
\label{eqn:Bayes_sym_knn_reg}
\end{align}
The symmetric $k$-NN regression estimate in Eq (\ref{eqn:sym_nn_reg_est}) can be described as follows.
\begin{align}
   m_n^{\mathrm{S{\it k}NNR}} (\vec x)=\frac{\sum_{i=1}^N w_{\mathrm{S}\it{k}\mathrm{NN}} (\vec x, \vec x_i) y_i }{\sum_{i=1}^N w_{\mathrm{S}\it{k}\mathrm{NN}} (\vec x, \vec x_i)}
\end{align}
We have two following theorems about the validity of the covariance matrix with $w_{\mathrm{S}\it{k}\mathrm{NN}}(\vec x_i,\vec x_j)$ and asymptotic property of the regression estimate. (See Appendix \ref{app1} and \ref{app2} for the proofs.)
\begin{theorem}
\label{theorem:sym_val}
   Covairance matrix $\tilde{\mat C}$ is valid for Gaussian processes if $\sigma^2>0$.
\end{theorem}

\begin{theorem}
\label{theorem:sym_cov}
  $\mu_{f_{\mathrm{new}},{\mathrm{S}{\it k}\mathrm{NN}}} (=-\frac{1}{\tilde{\kappa}} \tilde{\vec k}^T \vec y)$ converges to symmetric $k$-$\mathrm{NN}$ regression as $\sigma^2/\sigma_0$ approaches $0$.
\end{theorem}

\subsubsection{Hyperparameter Selection}

\label{subsubsec:hyper_sel}

We describe the hyperparameter selection method for Bayesian M$k$NN regression proposed in \cite{kim2016Bayes_kr}. It can be also used for the hyperparameter selection for Bayesian S$k$NN regression proposed in this paper. We have the set of hyperparameters is $\Theta=\{ k, \sigma_0, \sigma  \}$, where $k$ is a interger greater than $0$.  These sets of hyperparameters can be selected through the Bayesian evidence framework by maximizing the log of the marginal likelihood \cite{Gibbs97efficientimplementation} as follows.
\begin{align}
    \Theta^* =& \mbox{argmax}_{\Theta} \mathcal{L}(\Theta),
\end{align}
where 
\begin{align}
\mathcal{L}(\Theta) =& \log p(\vec y|\Theta) \\
            =& \log \{ |2\pi \mat C|^{-\frac{1}{2}} \exp (-\frac{1}{2}\vec y^T \mat C^{-1} \vec
y) \} \\
            =& \frac{1}{2} \log |\tilde{\mat C}| - \frac{1}{2} \vec y^T \tilde{\mat C} \vec
y - \frac{N}{2} \log 2\pi, \label{eqn:bkr_log_ev}
\end{align}
where $\tilde{\mat C}=\mat L + \sigma^2 \mat I$.

For the
continuous hyperparameters (e.g., $\sigma$, $\sigma_0$), the derivative can be used to
optimize $\mathcal{L}$ with respect to $\Theta$, where the
derivative of $\mathcal{L}$ with respect to $\theta$ is given by
\begin{align}
\frac{\partial \mathcal{L}}{\partial \theta} = \frac{1}{2}
\mathrm{trace}(\tilde{\mat C}^{-1} \frac{\partial \tilde{\mat C}}{\partial
\theta}) - \frac{1}{2}\vec y^T \frac{\partial \tilde{\mat C}}{\partial \theta} \vec y,
\end{align}
where $\tilde{\mat C}=\mat L + \sigma^2 \mat I$.
The discrete hyperparameter $k$ can be
selected based on the value of $\mathcal{L}$ as
\begin{align}
      K^*	&= \mbox{argmax}_k  \mathcal{L}(\{k,\sigma,\sigma_0\}).   
\end{align}
On the other hand, the posterior distributions of the
hyperparameters given the data can be inferred by the Bayesian method via
Markov Chain Monte Carlo methods
similarly to \cite{Neal1997RC_GP,Williams95gaussian}. And the regression estimate can be averaged
over the hyperparameters rather than obtained by one fixed set of hyperparameters. This would produce better results but cost more computational power. This approach has not been taken in this paper

\section{Mutual and Symmetric $k$-NN Classification and Bayesian Selection Methods for $k$} 

\label{sec:mut_sym_knn_cla}

\subsection{Mutual and Symmetric $k$-Nearest Neighbor Classification}

Let us assume we have the data set $\mathcal{D}_n=\{ (\vec x_1, y_1), \ldots, (\vec x_n, y_n) \}$, where $\vec x_i \in \mathbf{R}^d$ and $y_i \in \{ C_1,C_2,\ldots,C_J \}$.  We describe mutual and symmetric $k$-NN classfication methods with the notations $\mathcal{N}_k(\vec x)$, $\mathcal{N}'_k(\vec x_j),$ and $\mathcal{M}_k(\vec x)$ used to describe mutual and symmetric $k$-nearest neighbor regression in Section  \ref{subsubec:mut_sym_nnr}. The mutual $k$-NN classification method is described as
\begin{align}
m^{\mbox{M$k$NNC}}_n(\vec x) = C_{ \mathrm{argmax}_{c} | \{ \vec x_j \in \mathcal{M}_k(\vec x) | y_j = c \} | }.
\end{align}

Motivated by the symmetrised modelling used in \cite{Cucala08BayesiankNNcla}, we describe  the symmetric $k$-NN classification method as
\begin{align}
m^{\mbox{S$k$NNC}}_n(\vec x) = C_{ \mathrm{argmax}_{c} [
	 | \{ \vec x_j \in \mathcal{N}_k(\vec x) | y_j = c \} |
	+ | \{ \vec x \in \mathcal{N}'_k(\vec x_j) | y_j = c \} | ] }. 
\end{align}
It is trivial to show that the class label that the model in \cite{Cucala08BayesiankNNcla} estimates with the highest class probability
is the same as the one that the above method presents. The model proposed by \cite{Cucala08BayesiankNNcla} can be regarded as a full Bayesian model for the symmetric $k$-NN classification mentioned above.

\subsection{Bayesian Selection Methods for $k$ }

In Section \ref{subsec:Bayes_mut_sym_nnr} we described Bayesian mutual and symmetric $k$-NN regression methods with the selection schemes for the hyperparameters including $k$. We show that mutual and symmetric $k$-NN classification can be done with Bayesian mutual and symmetric $k$-NN regression methods, if the target values of the data set is encoded properly from class labels. We describe how it can be done for the cases of binary-class and multi-class (more than 2 classes) classification.

\subsubsection{Binary-class Classification}
\label{sec:binary_cla}

In case of the binary-class classification, we set a new training data set $\mathcal{D}^{\mathrm{CR}}_n=\{ (\vec x_1, y^{\mathrm{NE}}_{1}), \ldots, (\vec x_n, y^{\mathrm{NE}}_n) \}$ with new class label encodings, where
\begin{align}
	y^{\mathrm{NE}}_i  = \left\{ \begin{array}{ll}
 -1  & \mbox{if $y_i = C_1$} \\
  1   & \mbox{if $y_i = C_2$}. \end{array} \right.
\end{align}
Now given a new test data point $\vec x$ we apply Bayesian M$k$NN regression for the new training data set $\mathcal{D}^{\mathrm{CR}}_n$, and then we have M$k$NN classification method based on the result of Bayesian M$k$NN regression:
\begin{align}
    y_{\mathrm{new}}^{\mathrm{M}{\it k}\mathrm{NN},\mathrm{NE}} 
        &= \mathrm{sgn}(\mu_{f_{\mathrm{new},{\mathrm{M}{\it k}\mathrm{NN}}}}) \\
    &= \mathrm{sgn} ( \sum_{i=1}^N  \delta_{\vec x_{\mathrm{new}} \sim_k \vec x_i} \cdot \delta_{\vec x_i \sim_k \vec x_{\mathrm{new}}} y_i^{\mathrm{NE}} )  \\
        &= \mathrm{sgn} ( - | \{ \vec x_j \in \mathcal{M}_k(\vec x_{\mathrm{new}}) | y_j = C_1 \} |  \nonumber \\
             &+  | \{ \vec x_j \in \mathcal{M}_k(\vec x_{\mathrm{new}}) | y_j = C_2 \} |  ). 
\end{align}
It is also trivial to show that
\begin{align}
 y_{\mathrm{new}}^{\mathrm{M}{\it k}\mathrm{NN}} = 
  C_{(y_{\mathrm{new}}^{\mathrm{M}{\it k}\mathrm{NN},\mathrm{NE}} + 3)/2}
  = m^{\mbox{M$k$NNC}}_n(\vec x).
\end{align}
For the symmetric $k$-NN classification, 
we apply Bayesian S$k$NN regression for the new training data set $\mathcal{D}^{\mathrm{CR}}_n$, and then we have
S$k$NN classification method based on the result of Bayesian S$k$NN regression:
\begin{align}
	y_{\mathrm{new}}^{\mathrm{S}{\it k}\mathrm{NN},\mathrm{NE}} 
		&= \mathrm{sgn}(\mu_{f_{\mathrm{new},{\mathrm{S}{\it k}\mathrm{NN}}}})  \\
	&=  \mathrm{sgn} ( \sum_{i=1}^N  ( \delta_{\vec x_{\mathrm{new}} \sim_k \vec x_i}  + \delta_{\vec x_i \sim_k \vec x_{\mathrm{new}}} )  y_i^{\mathrm{NE}} ) \\	
      &= \mathrm{sgn}  [   -  [ | \{ \vec x_j \in \mathcal{N}_k(\vec x_{\mathrm{new}}) | y_j = C_1 \} | \nonumber \\
	&~ + | \{ \vec x_{\mathrm{new}} \in \mathcal{N}'_k(\vec x_j) | y_j = C_1 \} |  ]   \nonumber \\
	&+  [ | \{ \vec x_j \in \mathcal{N}_k(\vec x_{\mathrm{new}}) | y_j = C_2 \} | \nonumber \\
	&~ + | \{ \vec x_{\mathrm{new}} \in \mathcal{N}'_k(\vec x_j) | y_j = C_2 \} |  ]    ].
\end{align}
It is also trivial to show that
\begin{align}
 y_{\mathrm{new}}^{\mathrm{S}{\it k}\mathrm{NN}} = 
  C_{(y_{\mathrm{new}}^{\mathrm{S}{\it k}\mathrm{NN},\mathrm{NE}} + 3)/2} 
  = m^{\mbox{S$k$NNC}}_n(\vec x).
\end{align}
The hyperparameters including $k$ can be selected by the methods described in Section \ref{subsubsec:hyper_sel}.

\subsubsection{Multi-class Classification}
For the multi-class classification (with more than 2 classes), we present two kinds of methods.
First, we use the traditional formulation ({\it formulation I}) used in multi-class Gaussian process classification \cite{williams1998bayesian}. We consider Bayesian mutual and symmetric $k$-NN regression with $J$ outputs when we have $J$ classes. The outputs are expressed as $f^1, f^2, \ldots, f^J$.  We assume that the $Jn \times Jn$ covariance matrix 
of the prior of $\vec f$ is
\begin{align}
	\mat C = \left[ \begin{array}{llll}
        \mat C_{f^1}  & \vec 0 & \ldots & \vec 0 \\
        \vec 0 & \mat C_{f^1} & \ldots &  \vec 0 \\
        \vec 0 & \vec 0 & \ldots & \mat C_{f^J}   \end{array} \right],
\end{align}
with the covariance function $\mathrm{Cov}(f_i^j,f_k^l) = \delta(j,l) c (\vec x_i, \vec x_k)$. Then we have
\begin{align}
	{\mat C}&=(\mat D - \mat  W + \sigma^2 \mat I_{Jn\times Jn})^{-1} \\
	{\mat C_{f^l}}&=(\mat D_{f^l} - \mat  W_{f^l} + \sigma^2 \mat I_{n\times n})^{-1} \\
	\mat W &= \left[ \begin{array}{llll}
        \mat W_{f^1}  & \vec 0 & \ldots & \vec 0 \\
        \vec 0 & \mat W_{f^1} & \ldots &  \vec 0 \\
        \vec 0 & \vec 0 & \ldots & \mat W_{f^J}   \end{array} \right],
\end{align}
where $[\mat W_{f^l}]_{ij}=w_{\mathrm{M}\it{k}\mathrm{NN}}(\vec x_i,\vec x_j)$ for M$k$NN case, or $w_{\mathrm{S}\it{k}\mathrm{NN}}(\vec x_i,\vec x_j)$ for S$k$NN case. When we set a new encoding for a target value as
\begin{align}
	y^{\mathrm{NE}_2}_{il} &= \left\{ \begin{array}{ll}
         1 & \mbox{if $y_i=C_l$};\\
         0 & \mbox{otherwise,}\end{array} \right.   
\end{align}
given a new test data point $\vec x$ we have the predictive mean
\begin{align}
  \mu_{f^l_{\mathrm{new}},\mathrm{M}\it{k}\mathrm{NN}}
=&\frac{\sum_{i=1}^N w_{\mathrm{M}\it{k}\mathrm{NN}} (\vec x_{\mathrm{new}}, \vec x_i) y^{\mathrm{NE}_2}_{il} }
{\sum_{i=1}^N w_{\mathrm{M}\it{k}\mathrm{NN}} (\vec x_{\mathrm{new}}, \vec x_i)+\sigma^2}  \\
 =&\frac{\sum_{i=1}^N  \delta_{\vec x_{\mathrm{new}} \sim_k \vec x_i} \cdot \delta_{\vec x_i \sim_k \vec x_{\mathrm{new}}j} y^{\mathrm{NE}_2}_{il}  }
{\sum_{i=1}^N \delta_{\vec x_{\mathrm{new}} \sim_k \vec x_i} \cdot \delta_{\vec x_i \sim_k \vec x_{\mathrm{new}}} +\sigma^2/\sigma_0 }.
\end{align}
Then we have the classification method based on the result of multivariate Bayesian M$k$NN regression
\begin{align}
	y_{\mathrm{new}}^{\mathrm{M}{\it k}\mathrm{NN},\mathrm{MUL-I}}
	&~ = C_{\mathrm{argmax}_l \mu_{f^l_{\mathrm{new}},\mathrm{M}\it{k}\mathrm{NN}}	} \\
	&~ = C_{ \mathrm{argmax}_{l} \sum_{i=1}^N  \delta_{\vec x_{\mathrm{new}} \sim_k \vec x_i} \cdot \delta_{\vec x_i \sim_k \vec x_{\mathrm{new}}} y^{\mathrm{NE}_2}_{il} } \\
	&~ = C_{ \mathrm{argmax}_{c} | \{ \vec x_j \in \mathcal{M}_k(\vec x_{\mathrm{new}}) | y_j = c \} | } \\
	&~ = m^{\mbox{M$k$NNC}}_n(\vec x).
\end{align}
Similarly, for symmetric $k$-NN classification we have the classification method based on the result of multivariate Bayesian S$k$NN regression
\begin{align}
	&y_{\mathrm{new}}^{\mathrm{S}{\it k}\mathrm{NN},\mathrm{MUL-I}} \nonumber \\
	&= C_{\mathrm{argmax}_l \mu_{f^l_{\mathrm{new}},\mathrm{S}\it{k}\mathrm{NN}}	}   \\
	&= C_{ \mathrm{argmax}_{l}  \sum_{i=1}^N  ( \delta_{\vec x_{\mathrm{new}} \sim_k \vec x_i} + \delta_{\vec x_i \sim_k \vec x_{\mathrm{new}}} ) y^{\mathrm{NE}_2}_{il} }   \\
	&= C_{ \mathrm{argmax}_{c} [
	 | \{ \vec x_j \in \mathcal{N}_k(\vec x_{\mathrm{new}}) | y_j = c \} |
	+ | \{ \vec x_{\mathrm{new}} \in \mathcal{N}'_k(\vec x_j) | y_j = c \} | ] } \\
	&=m^{\mbox{S$k$NNC}}_n(\vec x) ,
\end{align}
where
\begin{align}
    \mu_{f^l_{\mathrm{new}},\mathrm{S}\it{k}\mathrm{NN}}
=&\frac{\sum_{i=1}^N w_{\mathrm{S}\it{k}\mathrm{NN}} (\vec x_{\mathrm{new}}, \vec x_i) y^{\mathrm{NE}_2}_{il} }
{\sum_{i=1}^N w_{\mathrm{S}\it{k}\mathrm{NN}} (\vec x_{\mathrm{new}}, \vec x_i)+\sigma^2} \nonumber \\
 =&\frac{\sum_{i=1}^N  ( \delta_{\vec x_{\mathrm{new}} \sim_k \vec x_i} + \delta_{\vec x_i \sim_k \vec x_{\mathrm{new}}j} ) y^{\mathrm{NE}_2}_{il}  }
{\sum_{i=1}^N ( \delta_{\vec x_{\mathrm{new}} \sim_k \vec x_i} + \delta_{\vec x_i \sim_k \vec x_{\mathrm{new}}} ) +\sigma^2/\sigma_0 }.
\end{align}

As in  \cite{kim2006emep} we use another formulation ({\it formulation II}) to avoid a redundancy in the traditional formulation pointed by \cite{Neal1997RC_GP}.  We use $J-1$ outputs only without redundancy, which are $(J-1)$ differences among $\{f^1, f^2, \ldots, f^J\}$. We define $g_i^{y_i,j}(=f_i^{y_i}-f_i^{j})$ for $j \neq y_i$. We set $\vec g_i$ to $[g_1^{y_1,1}, \ldots, g_1^{y_1,y_1-1}, g_1^{y_1,y_1+1},$ $\ldots, g_1^{y_1,J}]^T$, and set $\vec g$ to $[\vec g_1^T, \vec g_2^T, \ldots, \vec g_n^T]^T $. 
For $\vec g$ we have the $(J-1)n \times (J-1)n$ covariance marix $\vec C^{\mathrm{MUL}}$ with the covariance function
$\mathrm{Cov}(g_i^{y_i,j},g_k^{y_k,l}) = (\delta(y_i,y_k)-\delta(y_i,l)-\delta(y_k,j)+\delta(j,l))c (\vec x_i, \vec x_k)$ for $y_i \neq j$ and $y_k \neq l$. (For the derivation, see \cite{kim2006emep}.)

Given a new test data point $\vec x$, we have the multiple outputs $g_{\mathrm{new}}^{y_{\mathrm{new}},l}$ for $y_{\mathrm{new}} \neq l$. For the simplicity we try to get the estimates of $g_{\mathrm{new}}^{1,l}$ ($l\neq 1$).   For mutual $k$-NN classification, we have the predictive mean as in typical GP regression, as follows.
\begin{align}
 \mu_{g^{1,l}_{\mathrm{new},\mathrm{M}k\mathrm{NN}}}
 &=  \mu_{f^1_{\mathrm{new}},\mathrm{M}\it{k}\mathrm{NN}} -  \mu_{f^l_{\mathrm{new}},\mathrm{M}\it{k}\mathrm{NN}} \\
 &=  -\frac{1}{\kappa_{l,\mathrm{M}k\mathrm{NN}}} \vec k_{l,\mathrm{M}k\mathrm{NN}}^T \vec 1, 
\end{align}
where $\kappa_{l,\mathrm{M}k\mathrm{NN}}$ and $\vec k_{l,\mathrm{M}k\mathrm{NN}}$ are obtained from $\mat C^{\mathrm{MUL}}$ 
with the function $w_{\mathrm{M}\it{k}\mathrm{NN}}$ and $\mat C^{\mathrm{MUL}}_{J(n-1)\times J(n-1)+1}$ with one additional $g^{1,l}_{\mathrm{new},\mathrm{M}k\mathrm{NN}}$ \cite{Gibbs97efficientimplementation,Rasmussen:2005:GPM:1162254}. 
Based on  $\{ \mu_{g_{\mathrm{new},\mathrm{M}k\mathrm{NN}}^{1,l}} | l \neq 1  \}$, we have the classification method based on the result of multivariate Bayesian M$k$NN regression.
\begin{align}
    y_{\mathrm{new}}^{\mathrm{M}{\it k}\mathrm{NN},\mathrm{MUL-II}} 
    &= \left\{ \begin{array}{lll}
         C_1 & \mbox{if $\mu_{g^{1,l}_{\mathrm{new},\mathrm{M}k\mathrm{NN}}} > 0$} \\
            & \mbox{for all $l \neq 1$};\\
       C_{ \mathrm{argmin}_l ~ \mu_{g^{1,l}_{\mathrm{new},\mathrm{M}k\mathrm{NN}}} }
	& \mbox{otherwise}\end{array} \right.     \\
	    &=  C_{\mathrm{argmax}_l \mu_{f^l_{\mathrm{new}},\mathrm{M}\it{k}\mathrm{NN}}	} \\
   &= y_{\mathrm{new}}^{\mathrm{M}{\it k}\mathrm{NN},\mathrm{MUL-I}}  \label{eqn:f2_equi_f1_mknn}
\end{align}

Similarly, for symmetric $k$-NN classification we have the predictive mean as in typical GP regression, as follows.
\begin{align} 
 \mu_{g^{1,l}_{\mathrm{new},\mathrm{S}k\mathrm{NN}}}  
  &=  \mu_{f^1_{\mathrm{new}},\mathrm{S}\it{k}\mathrm{NN}} -  \mu_{f^l_{\mathrm{new}},\mathrm{S}\it{k}\mathrm{NN}} \\
 &=  -\frac{1}{\kappa_{l,\mathrm{S}k\mathrm{NN}}} \vec k_{l,\mathrm{S}k\mathrm{NN}}^T \vec 1, 
\end{align}
where $\kappa_{l,\mathrm{S}k\mathrm{NN}}$ and $\vec k_{l,\mathrm{S}k\mathrm{NN}}$ are obtained from $\vec C^{\mathrm{MUL}}$ 
with the function $w_{\mathrm{S}\it{k}\mathrm{NN}}$ and $\mat C^{\mathrm{MUL}}_{J(n-1)\times J(n-1)+1}$ with one additional $g^{1,l}_{\mathrm{new},\mathrm{S}k\mathrm{NN}}$ \cite{Gibbs97efficientimplementation,Rasmussen:2005:GPM:1162254}. 
Based on  $\{ \mu_{ g_{\mathrm{new},\mathrm{S}k\mathrm{NN}}^{1,l} } | l \neq 1  \}$, we have the classification method based on the result of multivariate Bayesian S$k$NN regression.
\begin{align}
    y_{\mathrm{new}}^{\mathrm{S}{\it k}\mathrm{NN},\mathrm{MUL-II}} 
    &= \left\{ \begin{array}{ll}
         C_1 & \mbox{if $g^{1,l}_{\mathrm{new},\mathrm{S}k\mathrm{NN}} > 0$}\\
       & \mbox{for all $l \neq 1$};  \\
       C_{ \mathrm{argmin}_l ~ \mu_{g^{1,l}_{\mathrm{new},\mathrm{S}k\mathrm{NN}}} }
	& \mbox{otherwise} \end{array} \right.    \\
    &=  C_{\mathrm{argmax}_l \mu_{f^l_{\mathrm{new}},\mathrm{S}\it{k}\mathrm{NN}}	} \\
     &= y_{\mathrm{new}}^{\mathrm{S}{\it k}\mathrm{NN},\mathrm{MUL-I}}  \label{eqn:f2_equi_f1_sknn}
\end{align}
This latter formulation ({\it formulation II}) exactly leads to the binary classification formulation described in Section \ref{sec:binary_cla}, when $J$ is 2.

As can be seen in  Eq (\ref{eqn:f2_equi_f1_mknn})  and Eq (\ref{eqn:f2_equi_f1_sknn}),  both the formulations produce the same classification results when they have the same hyperparameters. The hyperparameters including $k$ can be selected by the methods described in Section \ref{subsubsec:hyper_sel}.  However, the hyperparameters selected by the methods with the {\it formulation I and II}  can be different because they use different covariance matrixes in the marginal likelihood. (The former one uses the $Jn\times Jn$ covariance matrix and the latter one uses $(J-1)n \times (J-1)n$ covariance matrix.)

In the computer simulations even with the identical $k$ there can be cases where the classification results by M$k$NN (or S$k$NN), the ones based on Bayesian M$k$NN (or S$k$NN) regression methods with formulation I, and the ones by Bayesian M$k$NN (or S$k$NN) regression methods with formulation II are different. One of the reasons for that is that the matrix calculation is approximate.  Another reason is that they are different in the ways how dealing with vote tie cases.  When vote ties occur, in M$k$NN (or S$k$NN) the class label of the nearest neighbor among tied mutual neighbors (or among tied symmetric neighbors) is assigned. However, in the methods based on Bayesian M$k$NN (or S$k$NN) regression, the class label with the lowest index is assigned, because information on nearest neighbors are not available in themselves.

\section{Simulation Results}

\label{sec:sim_res}

\begin{figure}
\centering
\begin{tabular}{cc}
\includegraphics[width=4cm]{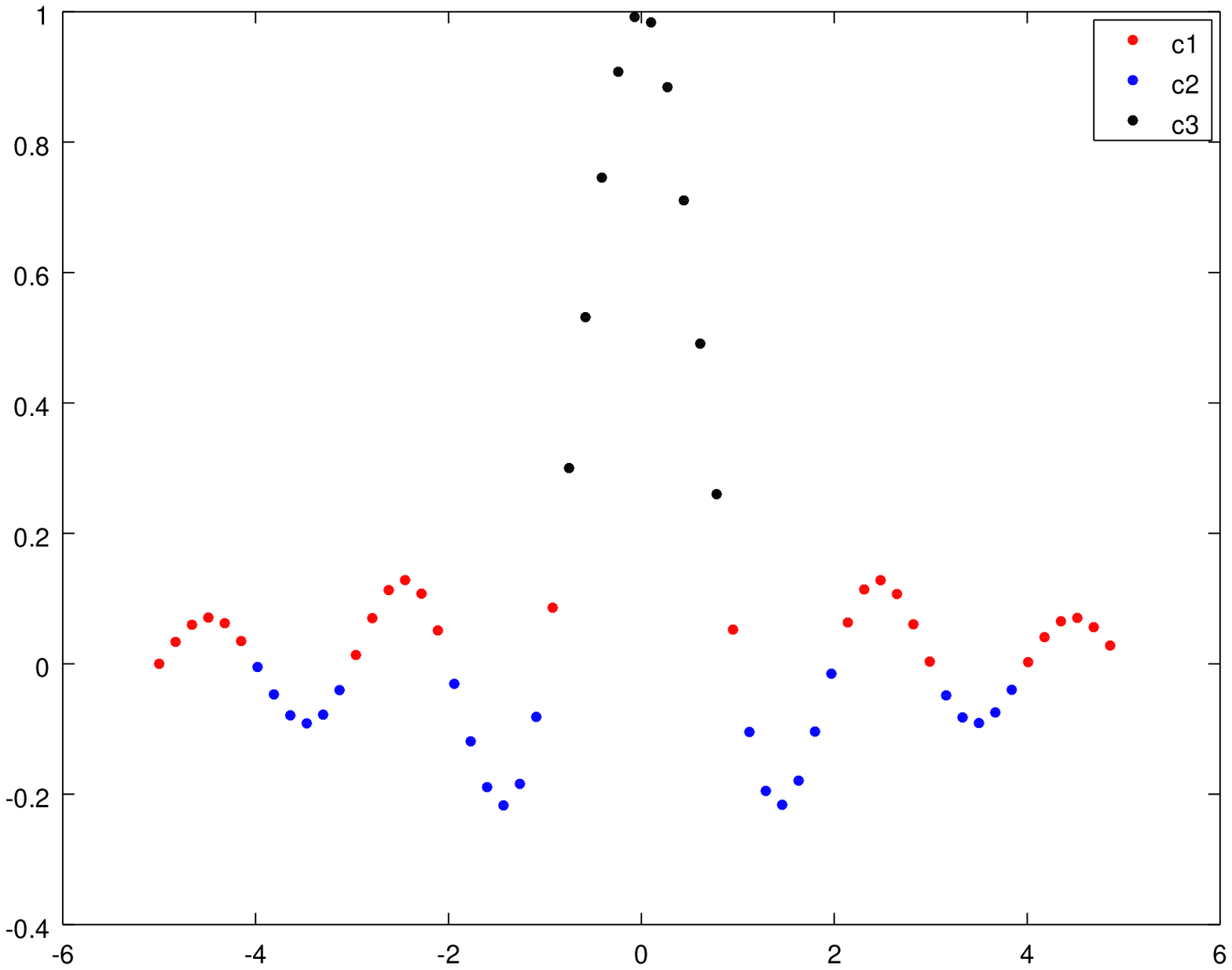} &
\includegraphics[width=4cm]{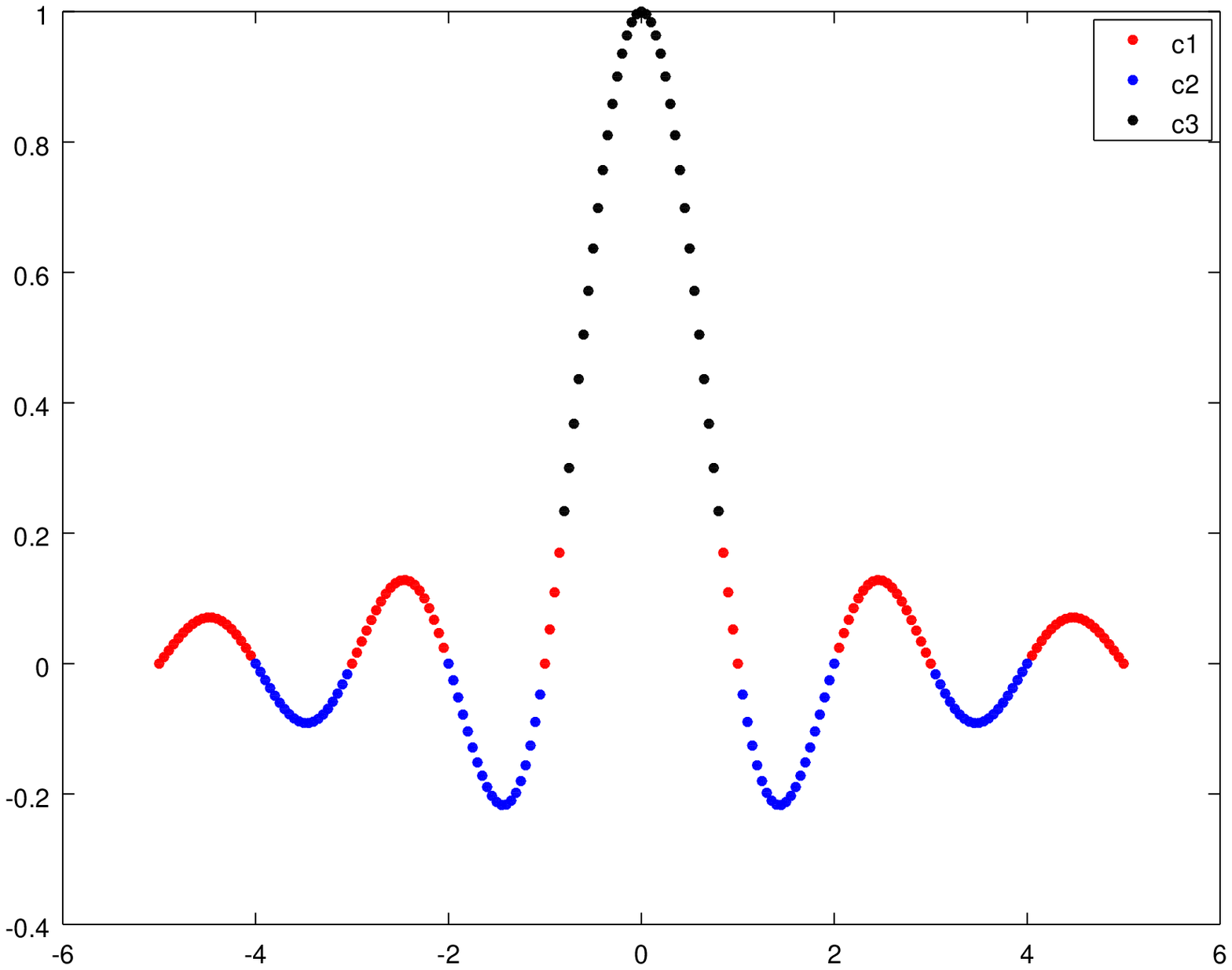} \\ 
(a) & (b)
\end{tabular}
\caption{ The plots of the Sinc3C data set: (a) the training set, (b) the test set}
\label{fig.1}
\end{figure}

To demonstrate the proposed methods first we did simulations for an artificial data set. To generate an artificial data set, we used the equation $\mathrm{sinc}(x) = \frac{\sin(\pi x)}{\pi x}$ for the sinc function. We took the points equally spaced with the interval 0.17  between -5 and 5. We assigned class labels 1, 2, 3 to those points according to intervals which the function values at those points belong to among $(-\infty, 0), [0, 0.2), [0.2, -\infty)$. We made up the training set with those points as inputs and with the assigned labels as target values. The data set is plotted in Figure \ref{fig.1}. We call this data set the Sinc3C data set. 

We applied M$k$NN and S$k$NN classification methods based on Bayesian M$k$NN and S$k$NN regression methods. We used both the {\it formulation I} requiring $J$ outputs and the {\it formulation II} requiring $J-1$ outputs. We tried the simulation repeatedly with different initial values for $\sigma_0,\sigma$, and found that one of the lowest marginal likelihoods is reached with the initial value 300, 3.  We also applied M$k$NN and S$k$NN classification methods with $k$ selected in the proposed methods, respectively, for each formuation. For comparison, we also applied $k$-NN\footnote{When vote ties occur, in $k$-NN classification the class label of the nearest neighbor among tied neighbors is assigned.}, M$k$NN, S$k$NN classification methods with the parameter $k$ selected by the leave-one-out cross-validation method.

\begin{figure}
\centering
\includegraphics[width=8cm]{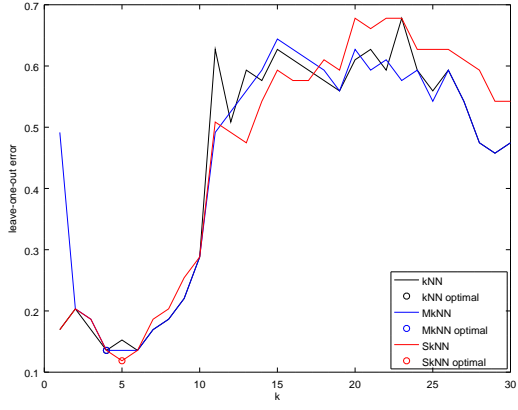}
\caption{ The leave-one-out errors of $k$-NN, mutual $k$-NN, symmetric $k$-NN classification according to $k$ for the Sinc3C training set. The points 'o' represent optimal ones.}
\label{fig.2}
\end{figure}

\begin{figure}
\centering
\begin{tabular}{c}
\includegraphics[width=8cm]{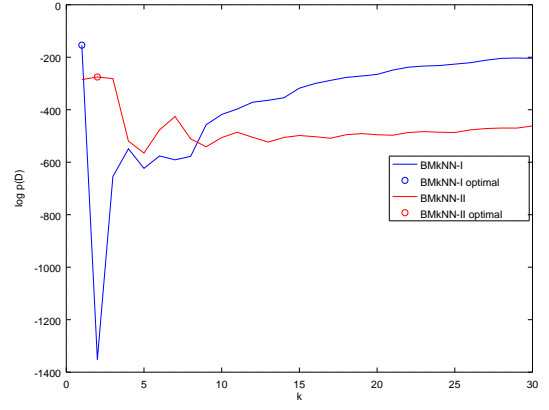} \\ 
(a) \\
\includegraphics[width=8cm]{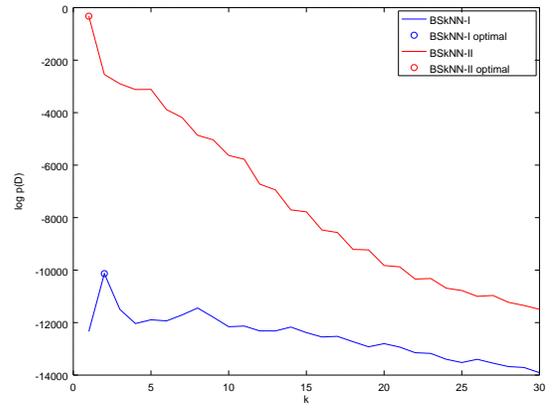} \\
(b)
\end{tabular}
\caption{ (a) $k$ vs. log evidence for BM$k$NN-I and BM$k$NN-II,
(b) $k$ vs. log evidence for BS$k$NN-I and BS$k$NN-II : The points 'o' represent the optimal ones}
\label{fig.3}
\end{figure}

Figure \ref{fig.2} shows the leave-out-errors of $k$-NN, M$k$NN, S$k$NN classification methods for the Sinc3C training set according to the parameter $k$. Figure \ref{fig.3} shows the log evidence of BM$k$NN, BS$k$NN regression models with the multi-class formulation I and II for the Sinc3C training set according to the parameter $k$. BM$k$NN-I and BS$k$NN-I represent M$k$NN and S$k$NN classification with the formulation I based on Bayesian M$k$NN and S$k$NN regression, respectively.  Likewise, BM$k$NN-II and BS$k$NN-II represent M$k$NN and S$k$NN classification with the formulation II based on Bayesian M$k$NN and S$k$NN regression, respectively. 

Table \ref{table:BkNNC_sinc} shows the classification error rates and $k$ selected for all the methods applied to the Sinc3C data set.  M$k$NN (B-I $k$) and S$k$NN (B-I $k$) represent M$k$NN and S$k$NN classification with the parameter $k$ selected in BM$k$NN-I, and BS$k$NN-I, respectively. 
M$k$NN (B-II $k$) and S$k$NN (B-II $k$) represent M$k$NN and S$k$NN classification with the parameter $k$ selected in BM$k$NN-II, and BS$k$NN-II, respectively. As can be seen in Table \ref{table:BkNNC_sinc}, BM$k$NN-II, BS$k$NN-II, M$k$NN (B-I $k$), and S$k$NN (B-II $k$) perform significantly better than all the other methods. 

\begin{table}[ht]
\caption{Classification error rates and the parameter $k$'s by various methods for the Sinc3C data set}
\centering 
\begin{tabular}{c c c} 
\hline\hline 
 Methods &  MSE  & $k$ selected \\ [0.5ex] 
\hline 
$k$-NN & 0.079602 & 2\\ 
M$k$NN  & 0.064677  & 3\\
S$k$NN  & 0.064677 & 5\\    
BM$k$NN-I  & 0.059701 & 1\\
BS$k$NN-I  &  0.089552 & 2\\
M$k$NN (B-I $k$) & {\bf 0.029851} & 1\\
S$k$NN (B-I $k$) & 0.074627 & 2\\
BM$k$NN-II  & {\bf 0.029851} & 2 \\
BS$k$NN-II  &  {\bf 0.029851} & 1\\
M$k$NN (B-II $k$) & 0.074627 & 2 \\
S$k$NN (B-II $k$) & {\bf 0.029851} & 1\\ 
[1ex]
\hline 
\end{tabular}
\label{table:BkNNC_sinc}
\end{table}

We applied the proposed methods and all the other methods to two real world data sets.  As the first real world data set, we use the Pima data set\footnote{Available from https://www.stats.ox.ac.uk/pub/PRNN/}. We used only the training set. It has 200 instances, 7 real-valued attributes, and 2 classes. We did 10 fold cross-validation to evaluate the performances of all the methods applied for the data set. The best results were obtained when the initial values of $(\sigma_0, \sigma^2)$ were set to $(1,10^{-6})$ for BS$k$NN, and when $(\sigma_0, \sigma^2)$ was fixed to $(150,1.5)$ for  BM$k$NN.

Table \ref{table:para_k_Pima} shows the parameter $k$'s selected by various methods such as cross-validation for $k$-NN, M$k$NN, S$k$NN classifications, and the proposed methods (binary-class case) for M$k$NN and S$k$NN classifications.  The abbreviation for the methods is the same as in Table \ref{table:BkNNC_sinc} except that this case has only the one (binary-class) formulation rather than two formulations.    Table \ref{table:BkNNC_Pima} shows the means and standard deviations of mean squared errors  (for 10 fold cross-validation) for the Pima data set. As can be seen in Table \ref{table:BkNNC_Pima}, M$k$NN and BM$k$NN perform better than all the other methods. 

\begin{table}[ht]
\caption{ The parameter $k$'s selected for each fold of the Pima data set by various methods: Cross-validation for $k$-NN, M$k$NN, S$k$NN classifications, and the proposed methods (binary-class cases) for M$k$NN and S$k$NN classifications}
\centering 
\tiny
\begin{tabular}{cccccccccccc} 
\hline\hline 
 Methods & f1 & f2 & f3 & f4 & f5 & f6 & f7 & f8 & f9 & f10 & Mean    \\ [0.5ex] 
\hline 
$k$-NN (CV)&   5  & 14 &  12 &  12 &  10  &  5 &  12 &  14 &  16 &   6 & 10.6 \\ 
M$k$NN (CV) & 49 &  27 &  75 &  73 &  78 &  80 &  79 &  70 &  84 &  69 & 68.4  \\
S$k$NN (CV)  &   51 &  48 &  47 &  53 &   5 &  36 &  36 &  84 &  11 &  53 & 42.4  \\    
BM$k$NN (binary)  &   93  & 108 &   97 &  101 &  100 &   98 &   94 &   96 &   83 &  97 & 96.7  \\
BS$k$NN (binary)  &  42 &  35 &  33 &  33 &  31 &  32 &  36 &  33 &  38 &  34 & 34.7 \\ [1ex]
\hline 
\end{tabular}
\label{table:para_k_Pima}
\end{table}

\begin{table}[ht]
\caption{Means and standard deviations of classification error rates of various methods for the Pima data set}
\centering 
\begin{tabular}{c c } 
\hline\hline 
 Methods &  MSE ($\mu\pm\sigma$)  \\ [0.5ex] 
\hline 
$k$-NN &  0.30500 $\pm$ 0.064334 \\  
M$k$NN  & {\bf 0.25000 $\pm$ 0.070711} \\ 
S$k$NN  & 0.26000 $\pm$ 0.077460 \\    
BM$k$NN (binary-class)  & {\bf 0.25000 $\pm$ 0.074536} \\
BS$k$NN (binary-class)  & 0.25500 $\pm$ 0.072457 \\
M$k$NN (B $k$) & 0.25500 $\pm$ 0.089598  \\ 
S$k$NN (B $k$) & 0.25500 $\pm$ 0.072457 \\ [1ex]
\hline 
\end{tabular}
\label{table:BkNNC_Pima}
\end{table}

To show how the methods work for the real world data set with more than two classes, we use New Thyriod data set \cite{Lichman:2013}. It has 215 instances, 5 real-valued attributes, and 3 classes. We did 10 fold cross-validation to evaluate the performances of all the methods applied for the data set. The best results were obtained when the initial values of $(\sigma_0, \sigma^2)$ were set to $(100,1)$, $(1,0.01)$, and $(100,1)$ for BS$k$NN-I, BM$k$NN-II, and BS$k$NN-II, respectively.  In case of BM$k$NN-I  $(\sigma_0, \sigma^2)$ were set and fixed to $(1, 0.0001)$.

Table \ref{table:para_k_NewT} shows the parameter $k$'s selected by various methods such as cross-validation for $k$-NN, M$k$NN, S$k$NN classifications, and the proposed methods (formulation I and II) for M$k$NN and S$k$NN classifications.  The abbreviation for the methods is the same as in Table \ref{table:BkNNC_sinc}.    Table \ref{table:BkNNC_NewT} shows the means and standard deviations of mean squared errors  (for 10 fold cross-validation) for the New Thyroid data set. As can be seen in Table \ref{table:BkNNC_NewT}, BS$k$NN-I perform better than all the other methods. 

\begin{table}[ht]
\caption{ The parameter $k$'s selected for each fold of the New Thyroid data set by various methods: Cross-validation for $k$-NN, M$k$NN, S$k$NN classifications, and the proposed methods (formulation I and II) for M$k$NN and S$k$NN classifications}
\centering 
\tiny
\begin{tabular}{c c c c c c c c c c c c } 
\hline\hline 
 Methods & f1 & f2 & f3 & f4 & f5 & f6 & f7 & f8 & f9 & f10 & Mean    \\ [0.5ex] 
\hline 
$k$-NN (CV)&   1  & 4 &  1 &  1 &  1 &  1 &  2 &  1 &  1 &  1 & 1.4 \\ 
M$k$NN (CV) &  42 &  39 &  43 &  27 &  25 &  4 &  60 &  24 &  43 &  35 & 34.2  \\
S$k$NN (CV)  &  1 &  5 & 2 &  2 &  2 &  2 &  2 &  2 & 2 &  2 & 2.2  \\    
BM$k$NN-I  &   22 &  22 &  24 &  19 &  22 &  18 &  19 &  19 &  22 &  22 & 20.9 \  \\
BS$k$NN-I  &   4 &  2 &  5 &  4 &  1 &  2 &  4 &  2 &  4 &  4 & 3.2 \\
BM$k$NN-II  & 10 &  10 &   8  & 10  & 10 &  11 &  10 &  10 &  10 &  10 & 9.9 \\
BS$k$NN-II  &  1  & 2 &  2 &  2 &  2 &  2 &  2 &  2 &  2 &  2 &  1.9 \\  [1ex]
\hline 
\end{tabular}
\label{table:para_k_NewT}
\end{table}

\begin{table}[ht]
\caption{Means and standard deviations of classification error rates of various methods for the New Thyroid data set}
\centering 
\begin{tabular}{c c } 
\hline\hline 
 Methods &  MSE ($\mu\pm\sigma$)  \\ [0.5ex] 
\hline 
$k$-NN & 0.041991 $\pm$   0.040510 \\   
M$k$NN  & 0.050866 $\pm$  0.062227  \\ 
S$k$NN  & 0.046537 $\pm$   0.053345 \\   
BM$k$NN-I  & 0.046320 $\pm$ 0.060985 \\
BS$k$NN-I  & {\bf 0.032684 $\pm$ 0.031835} \\
M$k$NN (B-I $k$) &0.050866 $\pm$ 0.066152 \\ 
S$k$NN (B-I $k$) & 0.037229 $\pm$ 0.047617  \\ 
BM$k$NN-II  & 0.037229 $\pm$ 0.053066  \\
BS$k$NN-II  &  0.042208 $\pm$ 0.046817 \\
M$k$NN (B-II $k$) & 0.046320 $\pm$  0.065328 \\ 
S$k$NN (B-II $k$) & 0.037446 $\pm$ 0.043098 \\ [1ex]
\hline 
\end{tabular}
\label{table:BkNNC_NewT}
\end{table}

\section{Conclusion}

We have proposed symmetric $k$-NN classification method, which is another variate of $k$-NN classification method.  We have proposed methods to select the parameter $k$ in mutual and symmetric $k$-NN classification methods. The selection problems boil down to the ones for the parameter $k$ in Bayesian mutual and symmetric $k$-NN regression methods, because Bayesian mutual and symmetric $k$-NN classifications can be done by Bayesian mutual and symmetric $k$-NN regression methods with new multiple-output encodings of target values.  For that purpose two kinds of encodings were proposed. The simulation results showed the proposed methods is comparable to or better than the selection by the leave-one-out cross-validation methods.

% Acknowledgements should go at the end, before appendices and references

%\acks{We would like to acknowledge . }


% or


\appendices
\section{Proof of Theorem \ref{theorem:sym_val}}
\label{app1}
Theorem  \ref{theorem:sym_val} can be done similarly to the proof of Theorem 1 in \cite{kim2016Bayes_kr} as follows. \\
    (1) Since Laplacian matrix $\mat L(=\mat D - \mat W)$ is positive semidefinite \cite{merris1994laplacian}, for $\sigma^2>0$ $\tilde{\mat C}(=\mat L + \sigma^2 \mat I)$ is positive definite. So $\tilde{\mat C}$ is positive definite.  \\
   (2) Since $\tilde{\mat C}^T = (\mat D - \mat W + \sigma^2 \mat I)^T
   		= \mat D^T - \mat W^T + \sigma^2 \mat I^T = \mat D - \mat W + \sigma^2 \mat I
	        = \tilde{\mat C}$,
	$\tilde{\mat C}$ is symmetric. \\
   From (1) \& (2), by Theorem 7.5 in \cite{stefanica2014linear} $\tilde{\mat C}$  is a valid covariance matrix. QED. 

\section{Appendix B. Proof of Theorem \ref{theorem:sym_cov}}
\label{app2}
  In case $\sum_{i=1}^N \{ \delta_{\vec x_j \sim_k \vec x_i} + \delta_{\vec x_i \sim_k \vec x_j} \} = 0$, it is trivial by Eq (\ref{eqn:sym_nn_reg_est}) and (\ref{eqn:Bayes_sym_knn_reg}). 

  Otherwise, take a small positive $\epsilon< m^{\mbox{S}k\mbox{NNR}}_n(\vec x) $ . \\
  Set $\delta = [ \sum_{i=1}^N \{  \delta_{\vec x_j \sim_k \vec x_i} + \delta_{\vec x_i \sim_k \vec x_j} \}  ] / \{ \frac{m^{\mbox{S}k\mbox{NNR}}_n(\vec x)}{\epsilon} -1 \}$.
  Then, 
    if $||\sigma^2/\sigma_0||<\delta $,
                $ || \mu_{\vec f_U,\mathrm{S}{\it k}\mathrm{NN}} - m^{\mbox{S}k\mbox{NNR}}_n(\vec x)  || < \epsilon $.      By the $(\epsilon,\delta)$ definition of the limit of a function, we get the statement in the theorem.
   QED.

%\section*{Acknowledgment}


\ifCLASSOPTIONcaptionsoff
  \newpage
\fi



%\bibliographystyle{IEEEtran}

\newpage

\bibliographystyle{IEEEtran}
\bibliography{paper}

%\begin{thebibliography}{1}
%\end{thebibliography}


%\end{IEEEbiography}
%% insert where needed to balance the two columns on the last page with
%%\newpage


%\vfill


\end{document}